\documentclass[runningheads]{llncs}
\usepackage{pdfpages}
\usepackage{graphicx}
\usepackage{amsmath}
\usepackage{multirow}
\begin{document}
\title{A Computer Vision Based Approach for Driver Distraction Recognition using Deep Learning and Genetic Algorithm Based Ensemble}
\titlerunning{Driver Distraction Recognition using GA Ensemble}

%
\author{Ashlesha Kumar\inst{1}\ \and
K.S. Sangwan\inst{2} \and
Dhiraj\inst{3}}
\authorrunning{A. Kumar et al.}
%
\institute{Department of Computer Science and Information Systems(CSIS), Birla Institute of Technology and Science(BITS) Pilani, Pilani Campus, India
\email{kumar.ashlesha@gmail.com}\\
\and
Department of Mechanical Engineering, Birla Institute of Technology and Science(BITS) Pilani, Pilani Campus, India\\
\email{kss@pilani.bits-pilani.ac.in}\\
\and
Central Electronics Engineering Research Institute (CSIR-CEERI), Pilani, India
\email{dhiraj@ceeri.res.in}
}
\maketitle              
\begin{abstract}
As the proportion of road accidents increases each year, driver distraction continues to be an important risk component in road traffic injuries and deaths. The distractions caused by increasing use of mobile phones and other wireless devices pose a potential risk to road safety. Our current study aims to aid the already existing techniques in driver posture recognition by improving the performance in the driver distraction classification problem. We present an approach using a genetic algorithm-based ensemble of six independent deep neural architectures, namely, AlexNet, VGG-16, EfficientNet B0, Vanilla CNN,Modified DenseNet and InceptionV3 + BiLSTM. We test it on two comprehensive datasets, the AUC Distracted Driver Dataset, on which our technique achieves an accuracy of 96.37\%, surpassing the previously obtained 95.98\%,and on the State Farm Driver Distraction Dataset, on which we attain an accuracy of 99.75\%. The 6-Model Ensemble gave an inference time of 0.024 seconds as measured on our machine with Ubuntu 20.04(64-bit) and GPU as GeForce GTX 1080.

\end{abstract}
\section{Introduction}
Road accidents are increasingly becoming one of the leading causes of fatalities, affecting both the developed and developing countries alike. The Global Status Report on Road Safety [1] revealed that such crashes and accidents claim roughly a million lives on a yearly basis. The count of people enduring disabling injuries as a result is even higher (ranging from approximately 20 to 50M). Most of these road traffic accidents are caused by drivers being distracted due to various factors such as conversations with co passengers or use of mobile devices. This leads to delayed response time and thus increases chances of accidents. Hence it is imperative to develop an accurate system which is capable of real-time detection of drivers distracted by various factors in their environment and so an alarm can be raised to alert them in time. Distracted driving, as defined by the NHTSA [2], is “any activity that diverts attention from driving”. The Center for Disease Control and Prevention (CDC) puts forward a more inclusive definition by bifurcating distractions into those caused visual, manual and cognitive sources. Cognitive distractions divert an individual’s mind away from the situation at present, manual distractions directly involve a person taking his hand off the steering wheel whereas visual distractions cause a person to take their eyes off the road.

Our research is focused on the manual category, targeting distractions of the form of talking or texting on phone, adjusting radio, fixing hair and makeup, eating or drinking and reaching behind to pick up stuff, sample images of which are depicted in Fig 1 and Fig 2. We propose a technique involving a genetically weighted ensemble of six end -to- end deep learning architectures consisting of convolutional neural networks and their combinations with recurrent neural networks which surpasses the state-of-the-art accuracies obtained on two of the most popular driver distraction datasets. We use AlexNet, VGG-16, InceptionV3 + BiLSTM, EfficientNet B0, Vanilla CNN and a modified hierarchical variant of DenseNet-201 as our individual branches in the ensemble. Our experiments are divided into two case studies, first on the AUC Distracted Driver (V1) Dataset results of which we compare with another study on GA weighted ensemble conducted by [3], and second on the State Farm Driver Distraction Dataset, where we evaluate our studies in comparison to three previous studies done on the same dataset.

The paper is arranged as follows: Section 2 talks about the related literatures available in the field of distracted driving detection, while Section 3 explains the structure and composition of the two datasets used. Section 4 discusses our proposed approach in length, explaining the details of various branches used to form the ensemble and the genetic algorithm adopted to construct the same. Section 5 presents the evaluation results obtained from our experiments and a comparative analysis with respect to the previous studies conducted on the datasets. Section 6 presents the conclusion of our paper along with the scope of future research based on it.
\section{Related Works}
Advances in technologies endorsed by fields such as Machine Learning and Deep learning have permitted researchers in the last two decades to come up with a myriad of distraction detection techniques, the earliest ones being dominated by simple classifiers such as SVM and Decision Trees. Research in the field of distracted driver detection can broadly be classified into categories of traditional machine learning approaches and modern deep learning solutions. Zhao et al. [4] proposed another inclusive dataset- the South East University (SEU) Driving Posture dataset, containing images of drivers taken in a side-view fashion, covering a broader range of activities such as talking on the cell- phone, driving safely, operating the lever and eating foodstuff. They adopted an approach of extracting features by making use of contourlet transform, assessing the performance using four classifiers, out of which the Random Forests classifier showed the best performance, achieving an accuracy of 90.63\%. Subsequently, Yan et al. [5] presented an approach to identify driving postures by making use of deep convolutional neural networks, attaining an accuracy of 99.78\% on the SEU dataset. Eraqi et al. [3] proposed a novel dataset, the AUC Distracted Driver dataset, comprising 10 classes of distractions.

\begin{figure}
\makebox[\textwidth][c]{\includegraphics[width = 1.5 \textwidth]{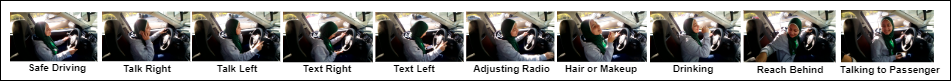}}
\caption{Ten classes of the AUC Distracted Driver(V1) Dataset}
\break

\makebox[\textwidth][c]{\includegraphics[width = 1.5 \textwidth]{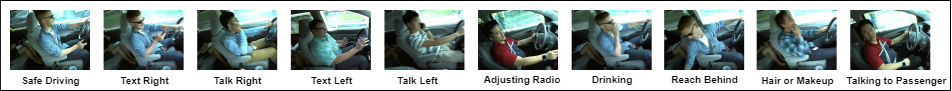}}
\caption{Ten classes of the State Farm Driver Distraction Dataset}
\end{figure}
\vspace{-3mm}
The inputs to the individual models in the ensemble were not same, with one of the networks being supplied with only raw images, another with skin-segmented images, while the remaining three networks were provided with images of hands, face and both hands-and-face respectively. Further progress in the field witnessed researchers and scientists combining both CNNs and RNNs to achieve better performance and accuracy in the detection of driver distraction. In 2019, Munif Alotaibi and Bandar Alotaibi [6] introduced a combination of three elaborate deep learning architectures. Their approach was tested on State Farm and AUC Datasets, yielding an excellent accuracy of 99.3\% on a fifty percent train/test split of the former one. Another benchmark was set by the Vanilla CNN architecture proposed by Jamsheed et al. [7] in 2020, giving an overall accuracy of 97\% on the State Farm Dataset. 
\section{Dataset Information}
\subsection{American University of Cairo (AUC) Distracted Driver (V1) Dataset}
The AUC Distracted Driver (V1) dataset [8] was the first publicly available dataset for distracted driver detection. Sample frames of the dataset are shown in Fig 1. It consists of participants from seven countries consisting of both males (22) and females (9). Videos were shot in 4 different cars: Proton Gen2 (26), Mitsubishi Lancer (2), Nissan Sunny (2), and KIA Carens (1)[8]. The dataset comprises of 12977 frames divided into 10 classes, as depicted by Table 1. 

\begin{table}[]
\begin{center}
\caption{Summary of the AUC Distracted Driver(V1) Dataset}\label{tab1}
\begin{tabular}{p{1cm}p{4cm}p{2cm}}
\hline
Class & Activity             & No. of Frames \\
\hline
0     & Safe Driving         & 2764          \\
1     & Phone Right          & 975           \\
2     & Phone Left           & 1020          \\
3     & Text Right           & 1480          \\
4     & Text Left            & 917           \\
5     & Adjusting Radio      & 915           \\
6     & Drinking             & 1209          \\
7     & Hair or Makeup       & 901           \\
8     & Reaching Behind      & 869           \\
9     & Talking to Passenger & 1927          \\
\hline
Total &                      & 12977     
\\
\hline
\end{tabular}
\end{center}
\end{table}
\vspace{-10mm}
\subsection{State Farm Driver Distraction Dataset}
To assess the strength and performance of our approach in a more generalized fashion, we also used another recent dataset as part of our second case study on driver distraction classification, the State Farm Driver Distraction dataset. Sample frames from the dataset are depicted in Fig 2. This dataset was released as part of State Farm’s Distracted Driver Detection competition [9] organized on Kaggle in 2016. 

\begin{table}[]

\begin{center}
\caption{Summary of the State Farm Dataset}\label{tab2}
\begin{tabular}{p{1cm}p{4cm}p{2cm}}
\hline
Class & Activity             & No.   of Frames \\
\hline
0     & Safe   Driving       & 2489            \\
1     & Text Right           & 2267            \\
2     & Talk   Right         & 2317            \\
3     & Text Left            & 2346            \\
4     & Talk   Left          & 2326            \\
5     & Adjusting Radio      & 2312            \\
6     & Drinking             & 2325            \\
7     & Reaching Behind      & 2002            \\
8     & Hair   or Makeup     & 1911            \\
9     & Talking to Passenger & 2129            \\
\hline
Total   &                      & 22424 \\
\hline
\end{tabular}
\end{center}
\end{table}

\vspace{-8mm}

Similar to the AUC dataset, it consists of images belonging to 10 classes, with postures ranging from safe driving to distractions such as texting on the phone while driving (using left or right hands), talking on the phone (using left or right hands), conversing with a co-passenger, operating the radio, reaching behind, adjusting hair or makeup, and drinking. The dataset thus consists of 22,424 images, distributed among the ten classes as shown in Table 2.

\section{Proposed Methodology}
Our approach takes advantage of a genetically weighted ensemble of convolutional neural networks, all of which were trained on raw images of the driver distraction datasets. The images were resized to a fixed size of 224 x 224. 
\begin{figure}
\makebox[\textwidth][c]{\includegraphics[width =1.2 \textwidth]{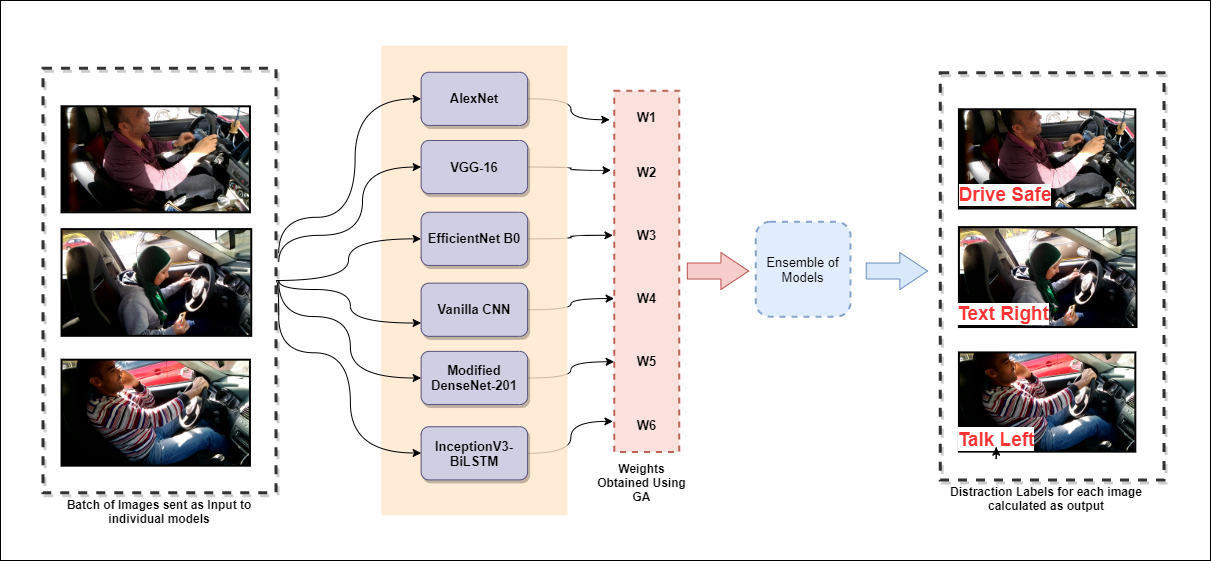}}
\caption{Summary of our proposed approach, employing a genetically weighted ensemble of net-works. The batch containing input images is fed to the networks, and the output produces labels for each of the individual images}
\end{figure}
\vspace{-4mm}
For our purpose, we conducted an exploratory analysis in which several branches were trained on each of the datasets separately, each branch having a different net-work architecture. We then calculate the weighted sum of the outputs of all classifiers employing a genetic algorithm, yielding the final class probability vector. The characteristics of each of the six branches are described in the following sections. The system overview is shown in Figure 3.t
\subsection{Independent Classifier Branches}
We trained several neural network architectures separately, adopting some of the finest ones previously achieving state-of-the-art performance on various computer vision tasks. 
\subsubsection{AlexNet} 
The AlexNet model was proposed in 2012 by Krizhevsky et al. [10]
This model clinched the first place in the 2012 ImageNet competition with a
top-5 error rate of 15.3\%. It has five convolutional layers, some of which are followed by max-pooling layers, and three fully-connected layers. ReLU activation
function is applied to the output feature maps of every convolutional layer. We trained our AlexNet models from scratch without any modifications to the original architecture, except the addi-tion of batch normalization before all ReLU
activations and the final SoftMax layer.We used an RMSProp optimizer and ran the model for 50 epochs with a batch
size of 32 images.

\subsubsection{VGG-16}VGG-16 was introduced in 2014 by Simonyan and Zisserman [11]. It is deeper than AlexNet, and consists of 13 convolutional layers followed by three fully connected layers. The input to the VGG ConvNets is a fixed-size 224 × 224 RGB image. The image is passed through a stack of convolutional layers, where filters are used with a very small receptive field: 3 × 3. The stack of convolutional layers is followed by three fully-Connected (FC) layers: the first two have 4096 channels each. The final layer is the soft-max layer.

For our purpose, we make use of a VGG-16 model pretrained on the ImageNet dataset as an integrated feature extractor, and the layers of the pre-trained model were frozen during training. An RMSProp optimizer with a learning rate of $
10^{-4}$ was used, keeping the number of epochs at 50 during the training process along with a batch size of 64 images.
\subsubsection{EfficientNet B0}Tan and Le [12] proposed a novel model scaling method in 2019 that uses a simple yet highly effective compound coefficient to scale up CNNs in a more structured manner. They developed a new baseline network, EfficientNet B0, by leveraging a multi-objective neural architecture search using the AutoML MNAS framework, that optimizes both accuracy and efficiency (FLOPS). As our third branch, we make use of the Efficient Net B0 model provided by Keras, pretrained on ImageNet Dataset. The pre-trained model is integrated into a new model, where two dense layers and a softmax activation are added to it, and the layers of the pre-trained model are trained along with the new model.
For the training process, we used an RMSProp optimizer, keeping the number of epochs as 30 and the batch size as 64 images.

\subsubsection{Vanilla CNN}Vanilla CNN architecture was proposed by Jamsheed et al in 2020[7], giving an overall accuracy of 97\% on the State Farm Dataset. The model is constructed with a total of 3 convolutional layers, a flatten layer and 3 dense layers. Activation function for all the convolutional layers and dense layers is ReLU, while the last layer em-ploys a softmax activation. Table 3 depicts the summary of the Vanilla CNN Model. This forms the fourth branch of our ensemble.
\begin{table}
\begin{center}
  \caption{Vanilla CNN Architecture}\label{tab1}  
\begin{tabular}{ p{1cm}p{4cm}p{4cm}}
\hline

Layer & Type                  & Output   Shape           
\\
\hline
1     & Convolutional   Layer & (None,   224, 224, 60) 
\\
2     & Max   Pooling Layer   & (None,   112, 112, 60  
\\
3     & Convolutional Layer   & (None, 112, 112, 90)   
\\
4     & Max Pooling Layer     & (None, 56, 56, 90)     
\\
5     & Convolutional   Layer & (None,   56, 56, 200)  
\\
6     & Max   Pooling Layer   & (None,   28, 28, 200)  
\\
7     & Flatten   Layer       & (None,   156800)      
\\
8     & Dense   Layer         & (None,   512)          
\\
9     & Dense   Layer         & (None,   128)          
\\
10    & Dense   Layer        & (None,   10)   
\\
\hline
\end{tabular}
\end{center}
\end{table}
\vspace{-4mm}
\subsubsection{Modified DenseNet-201 Hierarchical Model}Huang et al. [13] proposed an upgrade of the ResNet model, the DenseNet, which subsequently clinched the best paper award in CVPR2017. Verma et al. [14] proposed modified variants of DenseNet-201 employing hierarchical structures for posture detection on the Yoga-82 Dataset. Their main motive was to utilize hierarchy structure in the proposed dataset. One of the variants which we make use of consists of hierarchical connections built upon the DenseBlock 2 and DenseBlock 3 for three class levels. Level 1 originally consists of six classes, and level 2 is composed of 20 classes. The third level contains 82 classes, and the corresponding classification branch forms the main branch of the network.
\vspace{-4mm}
\subsubsection{InceptionV3 + BiLSTM}Inception architecture was first introduced in 2014 by Szegedy et al. [15]. It utilizes numerous kernel sizes in every convolutional layer to harness the power of varied kernel sizes, while at the same time prevents overfitting by avoiding deeper archi-tectures. LSTMs are an extension of recurrent neural networks, capable of learning long-term dependencies. LSTMs make use of the mechanism of cell states and its various gates to promote information flows, using which they can selectively choose to forget or remember things. The deep-bidirectional LSTMs [16,17] are an exten-sion of the described LSTM models in which two LSTMs are applied to the input data. 
A technique using pre-trained InceptionV3 CNNs [18] integrated with Bidirectional Long Short-Term Memory layers gave good results on the AUC Distracted Driver Dataset, outperforming other state-of-the-art CNN’s and RNN’s in terms of average loss and F1-score. We adopt this architecture as our sixth branch for the ensemble

For our purpose, we employ this variant with slight modifications, since the datasets used in our case studies are not explicitly hierarchical in nature. We set the class levels as 512, 128 and 10 respectively, and use the third variant proposed in [14] as our fifth branch for the main ensemble. A batch size of 64 images was used, and the network was trained for 30 epochs using an RMSProp optimizer.

\subsection{Genetic Algorithm(GA) Based Ensemble}
Our technique uses a classifier ensemble based on the idea that ensembles can serve as strong classifiers or a more accurate mechanism for prediction than the individual weak classifiers they consist of. Suppose that there are N such individual classifiers, $   C_{1}, C_{2}... C_{N} $. Solving a prediction problem involving m classes $ M_{1}, M_{2}...M_{m}$. Let us assume that depending on the classification problem, the features used and the training set, classifier $C_{1} $ is more efficient in classifying $
 M_{1}, C_{2}$ is more accurate on another class, say $
  M_{2}  or M_{3} $ and so on. If we form an ensemble of classifiers, $ C_{1}, C_{2}... C_{N} $ we cannot assign equal weights to all of them and hence a method has to be employed to determine the weight of vote given to each model. We adopt the genetic algorithm[19], a search heuristic that automatically and effectively finds the proper weights of all the eligible models. If the individual probability vectors are designated by $ V_{1} ...V_{N},$ produced as the output of the last softmax layer of each model, then in a weighted voting system, the final predictions are calculated in equation (1) as:
  \begin{equation}
      V_{final} = \frac{\sum_{i=1}^{N} w_{i}*V_{i}}{\sum_{i=1}^{N} w_{i}}
  \end{equation}
Where each chromosome consists of N genes, 
$w_1, w_2…w_N$. We adopt the Mean Squared Error (MSE) as our objective function and create an initial population of 48 individuals. We run the algorithm for 30 generations. The chromosome with the highest fitness score is selected in the end. 
\section{Experiments and Results}
As a first step, we constructed a GA weighted ensemble of all the six branches taken together. We conducted our experiments on the AUC Driver Distraction Dataset as our first case study using all the six models as an ensemble, and then on the State Farm Dataset as our second case study to further assess the strength of our approach by comparing it with the existing state-of-the-art obtained for the same.The training plots for all the individual branches depicting the training accuracy and training loss as a function of the number of epochs are shown in Figure 4. 
\subsection{Results on AUC Driver Distraction(V1) Dataset}
The results of all the individual branches trained on the AUC Dataset are shown in Table 4. 
\begin{table}[]
\begin{center}
\caption{Results obtained on Individual Branches for AUC Dataset}
\begin{tabular}{p{4cm}p{2cm}}
\hline
Model                   & Accuracy \\
\hline
AlexNet                 & 96.24\%  \\
InceptionV3-BiLSTM      & 95.28\%  \\
VGG-16                  & 95.3\%   \\
Modified   DenseNet-201 & 94.42\%  \\
Vanilla   CNN           & 95.76\%  \\

EfficientNet   B0       & 95.3\%  \\
\hline
\end{tabular}
\end{center}
\end{table}
\vspace{-5mm}
\begin{figure}

\makebox[\textwidth][c]{\includegraphics[width =1.2 \textwidth, height=14cm]{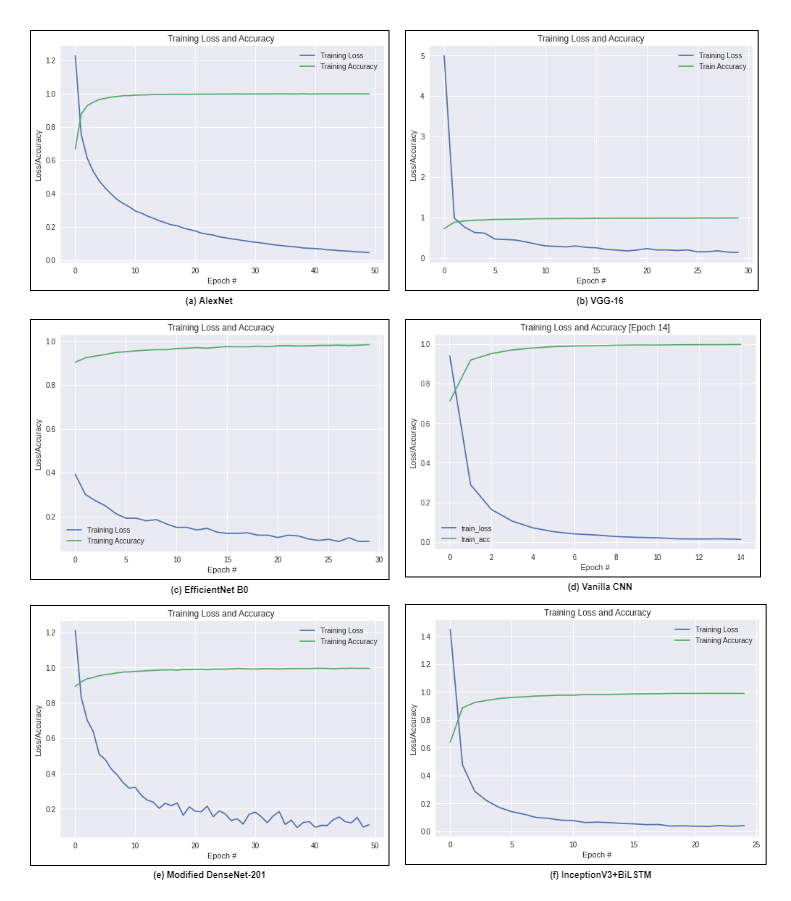}}
\caption{Training Curves of each of the individual architectural branches}

\makebox[\textwidth][c]{\includegraphics[width =1.2 \textwidth, height=7cm]{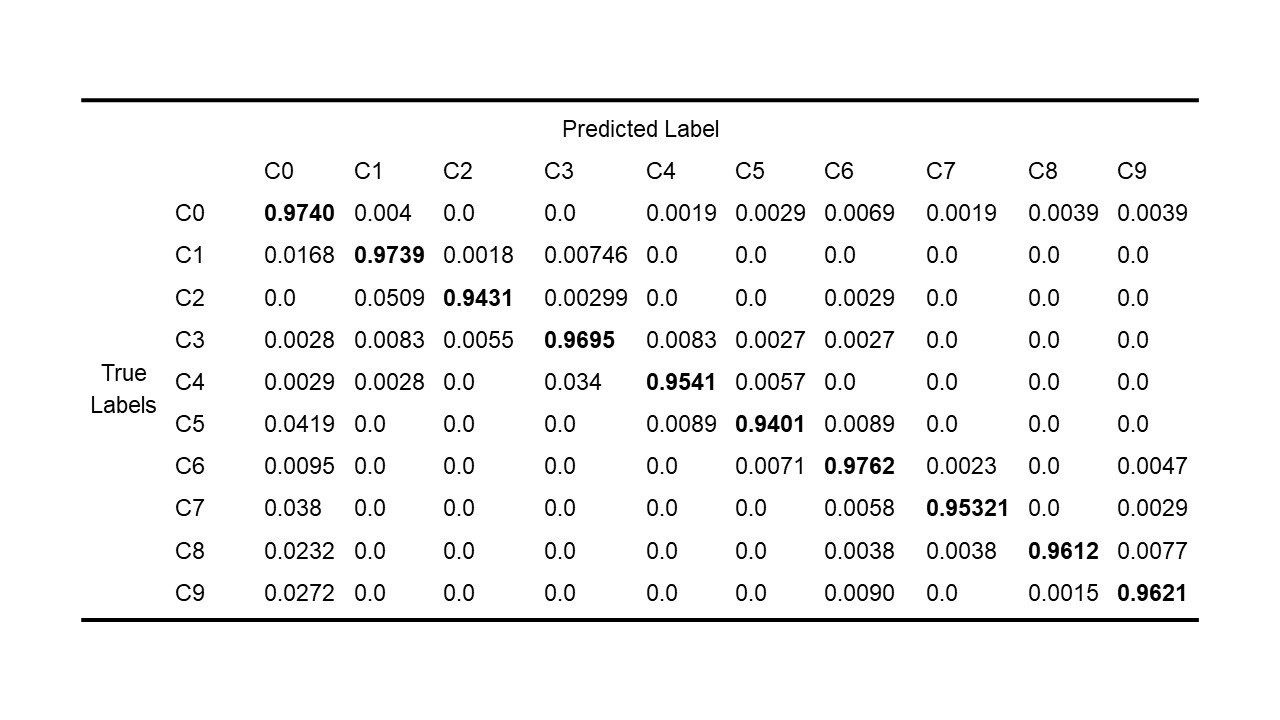}}
\vspace{-15mm}
\caption{Confusion Matrix for the GA Ensemble formed using the six branches stated earlier}

\end{figure}

\subsubsection{6-Model Ensemble}Our 6-Model GA-Weighted Ensemble gave an overall accuracy of 96.37\%, which was greater than the accuracies of all the individual branches used . This particular combination also showed marked improvement in the classification of the classes Safe Driving (2.06\%), Phone Right (0.76\%), Text Right (0.80\%) and Reaching Behind (3.36\%). involve the use of mobile phones by the same hand.

The confusion matrix shown in Figure 5 reveals that there is a confusion between the classes Phone Right and Text Right and between the classes Phone Left and Text Left. A possible reason could be because both actions involve the use of mobile phones by the same hand (right or left respectively). Another observation drawn was that “Hair and Makeup” was majorly confused for the “Talking to Passenger” positions, due to the striking similarity in the two positions, as depicted in Figure 6.

\begin{figure}
\makebox[\textwidth][c]{\includegraphics[width =1.2 \textwidth]{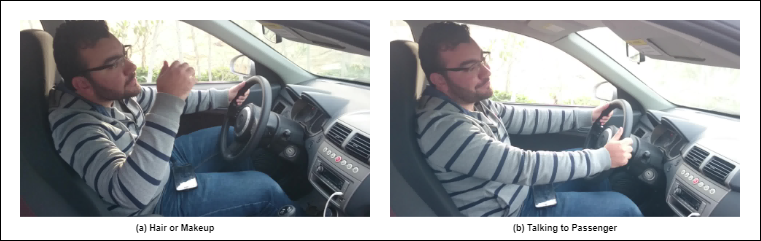}}
\caption{Comparison of “Hair or Makeup” and “Talking to Passenger” postures. The similarity in the positions cause the driver to bend towards the other side in both cases with movement of the right hand, hence causing the confusion in the decision making process of the network}
\end{figure}
\vspace{-7mm}
\subsection{Results on State Farm Driver Distraction Dataset}
We tested the ensemble on the State Farm Dataset and compared the results with three other studies conducted on the same dataset, [6], [7] and [20] using similar train/test partitions employed by the original authors, as depicted in Table 5.

Munif Alotaibi and Bandar Alotaibi [6] experimented with three different train/test partitions, achieving the highest accuracy of 99.3\% on the fifty percent train/test split. Our 6-Model ensemble exceeded the benchmark accuracy on almost all of the partitions, except one where it gave a similar performance. The comparison for all the references is represented by Table 5.

 Vanilla CNN [7] architecture was originally tested on the State Farm Dataset, and since our ensemble employed it as one of the branches, we used similar train/dev/test splits to compare our results with it as well. Our proposed method gave an accuracy of 99.57\% as compared to 97.66\% previously attained by [7]. This revealed that using a combination of classifiers gave better results as compared to using only a single model.
 
 K.R Dhakate and R. Dash[19] made use of a stacking ensemble technique on a different split of the State Farm Dataset, on which our technique achieves an accuracy gain of 2.75\%.
\vspace{-7mm}
 \begin{table}
 \caption{Results obtained on Different Train/Test Partitions of the State Farm Dataset and its performance comparison with other published results}
  \makebox[\textwidth][c]{
\begin{tabular}{|c|c c c | c c c|}
\hline
\multirow{2}{*}{Reference} & \multicolumn{3}{|c|}{Data Split}            & \multicolumn{3}{c|}{Accuracy}                                       \\
                           & Train Data \% & Val Data \% & Test data \% & Obtained  by  Authors & Our Ensemble & Accuracy Gain \\\hline
\multirow{3}{*}{{[}6{]}} & 10\%                     & -    & 90\% & 96.23   & 97.16\%   & 0.93\% \\
                         & 30\%                     & -    & 70\% & 98.92   & 98.923 \% & 0\%    \\
                         & 50\%                     &      & 50\% & 99.3    & 99.42\%   & 0.12\% \\
{[}20{]}                 & 64\% (13K Images taken)  & 18\% & 18\% & 97\%    & 99.75\%   & 2.75\% \\
{[}7{]}                  & 80\% (Total 17939 taken) & -    & 20\% & 97.66\% & 99.57\%   & 1.91\%\\
\hline
\end{tabular}
}

\end{table}
\vspace{-3mm}
\section{Conclusion}
As the use of mobile phones and technological devices increases at an aggressive pace, potential risk to road safety due to the distractions caused by these is also in-creasing exponentially. We present an approach that incorporates the strength of a number of advanced deep learning architectures to accurately predict whether a driver is distracted or driving safely. We make use of the nature inspired “genetic algorithm” to efficiently create a weighted ensemble. Our approach has been tested on two comprehensive datasets, and shows excellent results in terms of accuracy based performance gain on both of them with respect to already published results in literature. As a future work, the GA ensemble can be deployed on an embedded device and its performance can be evaluated in real-time. Research also needs to be done on how to effectively reduce the confusion between some of the classes (i.e., “Hair and Makeup” and “Talking to Passenger”, or “Phone Right and Text Right”) showing similar trends in most of the experiments. The network combination can also be used for generating labels for the unlabeled images in State Farm Dataset for further detailed studies.

%
%
%
%

\end{document}